\documentclass[10pt]{aphsc_article}    

\usepackage{cite} 
\usepackage{url}  
\usepackage[utf8]{inputenc} 
\usepackage{paratype}
\usepackage{graphicx}
\usepackage{lastpage}
\usepackage{booktabs}
\usepackage{amsmath}
\usepackage{color}
\graphicspath{{/}}

\usepackage{fancyhdr}
\usepackage[paperwidth=15cm,paperheight=21cm]{geometry}

\usepackage{float}
\begin{document}

\title{Some Reflections on the Set-based and the Conditional-based Interpretations of Statements in Syllogistic Reasoning}

\author{M. Pereira-Fari\~na\correspondingauthor$^1$%
         \email{M. Pereira-Fari\~na\correspondingauthor - martin.pereira@usc.es}
      }

\address{%
    \iid(1)Centro Singular de Investigaci{\'o}n en Tecnolox{\'i}as da Informaci{\'o}n (CiTIUS), Universidade de Santiago de Compostela, Santiago de Compostela, Spain
}%

\maketitle

\begin{abstract}
Two interpretations about syllogistic statements are described in this paper. One is the so-called set-based interpretation, which assumes that quantified statements and syllogisms talk about quantity-relationships between sets. The other one, the so-called conditional interpretation, assumes that quantified propositions talk about conditional propositions and how strong are the links between the antecedent and the consequent. Both interpretations are compared attending to three different questions (existential import, singular statements and non-proportional quantifiers) from the point of view of their impact on the further development of this type of reasoning.
\\
\\
{\footnotesize Submitted: 00/00/00. Accepted and Published: 00/00/00.}
\end{abstract}

\newpage

\section{Introduction}
\label{sc:Introduction}
A syllogism is defined as a deductive argument based on the chaining of terms using quantified statements. In its classical version~\cite{Aristotle1949}, it is composed by three propositions and a single conclusions is necessarily inferred from two premises. The statements are assertions with the form \emph{Q S are P}, where \emph{Q} is one of the four classical logic quantifiers (``all'', ``no'', ``some'' and ``some\dots not''), \emph{S} is the subject-term and \emph{P} is the predicate-term. All syllogisms involve three terms: the so-called \emph{extremes} (Minor and Major terms) constitute the conclusion and also appear in the premises; and the so-called \emph{middle term}, which only appears in the premises and links the extreme ones. The position of the extreme terms in the conclusion is fixed (the minor one is the subject-term and the major one is the predicate one) but in the premises their position can vary, as well as the one of the middle term. Like this, the four Aristotelian figures~\cite{
Aristotle1949} are generated, that are the four different classical syllogistic inference patterns.\par

Perhaps, the most famous example of syllogism is the one shown in Table~\ref{tab:AristotelianSyllogism}, where from the links between the terms \emph{human (beings)} and \emph{mortal beings} and between \emph{Greeks} and \emph{human beings}, the link between \emph{Greeks} and \emph{mortal (beings)} of the conclusion is inferred.

\begin{table}[h]%
	\centering
	\begin{tabular}{c}%
	  All human beings are mortal\\
	  All Greeks are human beings\\\hline
	  All Greeks are mortal
	\end{tabular}
	\caption{\label{tab:AristotelianSyllogism} Aristotelian Syllogism.}
\end{table}

Although Aristotelian Syllogistics was substituted by mathematical logic at the beginning of twentieth century~\cite{Kneale1968}, in its second half it was partially recovered~\cite{Lukasiewicz1957}. One of the aims of the resurgence of syllogistic studies was to improve the expressiveness capability of Aristotle's Syllogistics. During 1980s and early 1990s, different models that improve syllogistic reasoning appeared in the literature. We classify them into two categories: i) proposals that preserve Aristotelian knowledge representation and inference patterns; ii) proposals that explore alternative models for representing quantified statements and the inference process. 

\begin{figure}[h]
	\centerline{\includegraphics[keepaspectratio=true,width=0.60\textwidth]{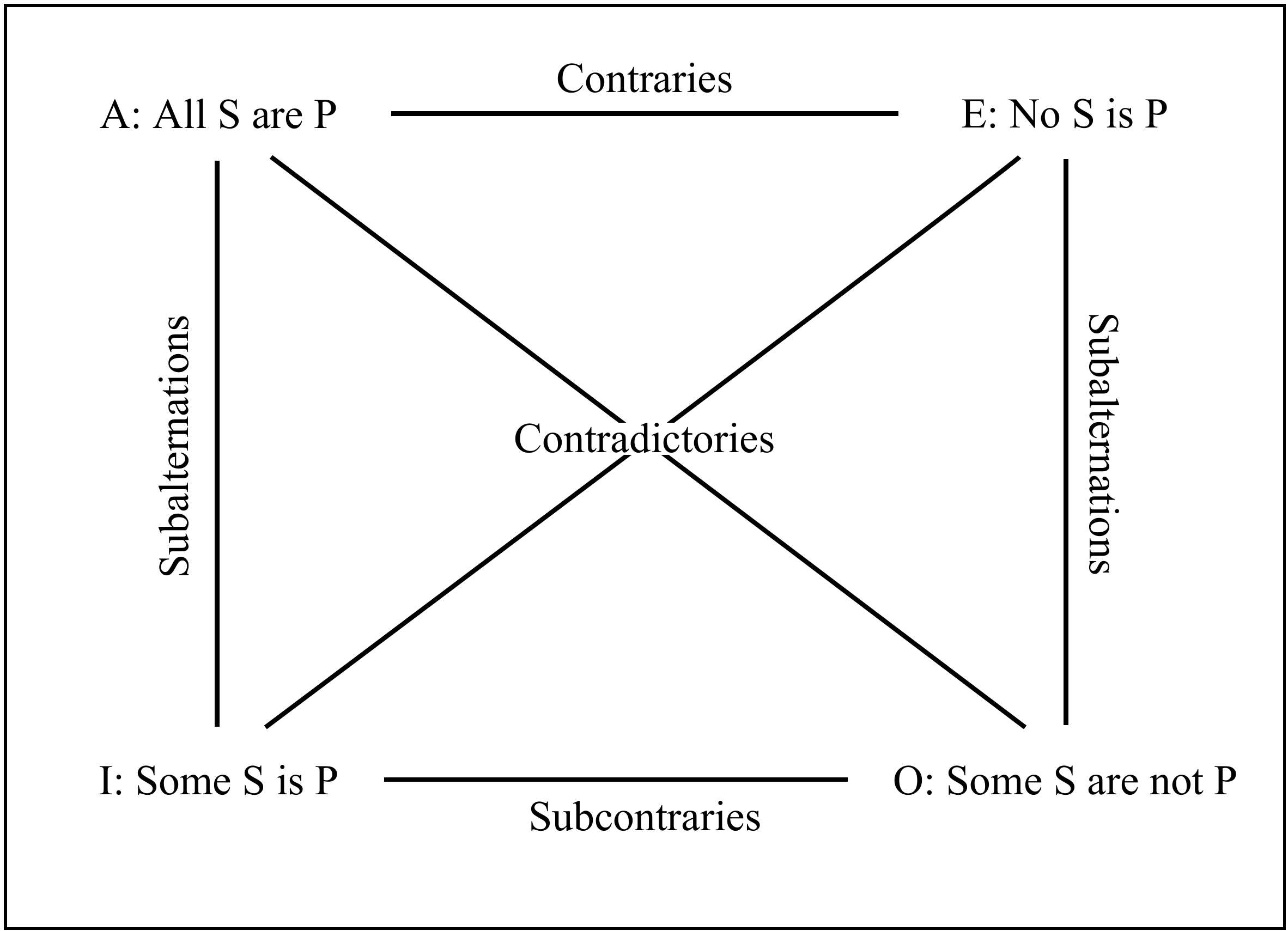}}
	\caption{Classical LSO.}
	\label{fig:LSOClassical}
\end{figure}

In the models of category i), we can find two different approaches. The first one, that involves different proposals~\cite{Peterson2000,Thompson1982,Murphree1997,Murinova2012}, proposes to improve Aristotelian Syllogistics introducing new quantifiers such as ``few'', ``most'', ``many'', etc. The core of Aristotelian Syllogistics is the Logic Square of Opposition (LSO) (see Fig.~\ref{fig:LSOClassical}). It is a square diagram where the logical relationships between the four type of Aristotelian syllogistic statements (corresponding to each one of the quantifiers) define the inference rules used by Aristotle for proving the validity of syllogisms. Thus, these new quantifiers are introducing between the universal (``all'', ``no'') and the existential (``some'', ``some\dots not'') ones preserving the Aristotle's inference rules. Some of these approaches use crisp definition~\cite{Peterson2000,Thompson1982,Murphree1997} for the quantifiers and others, fuzzy ones~\cite{Murinova2012}.\par 

The second approach of category i) assumes a different treatment for quantification. Fuzzy Syllogistics~\cite{Zadeh1985} interprets quantifiers as fuzzy numbers in the interval $[0,1]$. Thus, ``almost all'' is interpreted as the fuzzy number \emph{approximate 1} ($\tilde{1}$ using the typical fuzzy notation), which can correspond, for instance, with the fuzzy set $[0.95,0.97,0.98,1]$; ``few'', for instance, with the fuzzy number \emph{no more than 20\%}; $[0,0.8,0.12,0.2]$. Interval Syllogistics~\cite{Dubois1990}, on the other hand, interprets quantifiers in terms of intervals and, thus, ``almost all'' means, for instance, $[0.95,1]$ and ``few'' $[0,0.15]$. Interval Syllogistics can be interpreted as an intermediate step between crisp approaches a fuzzy ones.\par

In the models of category ii), it is proposed that quantified statements have a deep logical form which corresponds with a conditional proposition; e.g., ``all human beings are mortal'' means ``if \emph{x} is a human being, then \emph{x} is mortal'', where the the antecedent is the subject-term, the consequent is the predicate-term and the quantifier denotes how strong is the link between them. Thus, for the quantifier ``all'' the strength of the link is total while for ``some'' the link is very weak. In the literature, we can find different alternatives for modelling this link; for instance, Probabilistic Syllogistics~\cite{Oaksford2007}, that uses probability, or Support Logic Syllogistics~\cite{Spies1989}, that uses Dempster-Shafer Theory of Evidence.\par

Both categories, i) and ii), lead us to consider that two different interpretations about quantifies statements~\cite{Pereira2013}:
\begin{itemize}
	\item \textbf{Set-based interpretation}: They are the models of category i). This interpretation assumes that the syntactical form of the quantified statements is their logical form. The terms that constitute the propositions are sets (i.e., collection of elements) and the quantifier denotes a quantity-relationship between these sets. They are assertions (that can be true or false) that entails commitment with reality. For instance, ``all human beings are mortal'' means that the subject-term, the set of \emph{human beings}, is a subset of the predicate-set, the set of \emph{mortal beings}; ``no human being is a Martian'' means that the intersection between the set of \emph{human beings} and \emph{Martians} is empty. In this interpretation, the cardinality of the sets has a fundamental role and, therefore, a syllogism can be seen as an inference attending to the cardinalities between sets.
	\item \textbf{Conditional interpretation}: They are the models of category ii). This interpretation assumes that the syntactical form of the statement hides its genuine logical form: a conditional structure. Thus, the subject-term is the antecedent, the predicate-term is the consequent and the quantifier denotes how strong is the relationship between them. While set-based interpretation focuses on true descriptions in terms of sets, conditional one focuses on fix the truth-conditions for verifying the proposition. For instance, the sentence ``all human beings are mortal'' means ``if something is a human being, then it is mortal with the maximum strength in the link'' or even ``if \emph{any element of the universe has the property of being a human being}, then \emph{this element has the property of being a mortal being} with the \emph{maximum strength} in the link''. In this case, syllogisms show an inference schema closer to \emph{Modus Ponens}, where the truth of the antecedent allows us to conclude the 
consequent rather than a calculation of cardinalities between sets.
\end{itemize}

Both interpretations generate different approaches for dealing with syllogisms which are not always equivalent~\cite{Spies1989}. Thus, an analysis about the characteristics and presuppositions that each of these interpretations entails is a relevant task to be addressed for a further development of syllogistic reasoning. This is, precisely, the main objective of this paper.

In~\cite{Oaksford2007}, we can find a first discussion about this question. The probabilistic approach is defended as the best one as it is supported on experiments from psychology of reasoning. These show that human beings have a Bayesian rationality and all those quantifiers that are relevant from the point of view of reasoning using the conditional interpretation and probabilities for interpreting quantifiers.\par

Nevertheless, from the point of view of a computational approach, psychological experiments are not a strong enough proof. In this paper, we propose a new point of view to stimulate this discussion about which is the best interpretation for improving the current models of syllogistic reasoning and their implementation.

%


To achieve the aims explained before this paper is organized in the following sections: section~\ref{sc:SetBasedInterpretation} describes the main characteristics of set-based interpretation and some open questions to be considered; section~\ref{sc:ConditionalInterpretation} explains the model of syllogism performed by Probabilistic Syllogistics and its corresponding challenge questions; and, finally, in section~\ref{sc:Conclusions}, the main conclusions of this paper are exposed.

\section{Set-Based Interpretation: Models and Questions}
\label{sc:SetBasedInterpretation}
\begin{figure}[h]
	\begin{center}
	\includegraphics[keepaspectratio=true,width=0.45\textwidth]{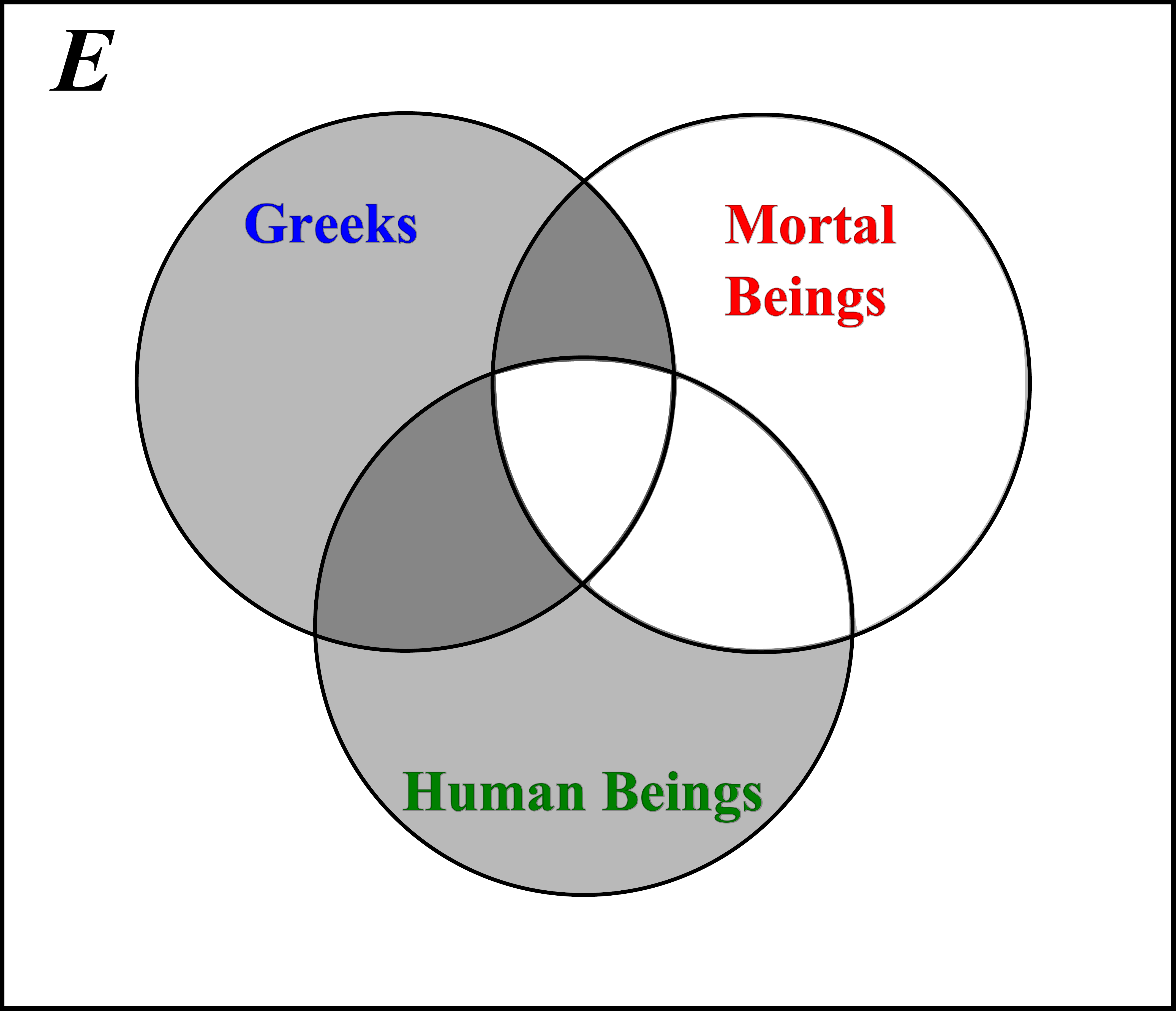}
	\caption{\label{fig:VennDiagramSyllogism} Venn Diagrams for the Aristotelian Syllogism of Table~\ref{tab:AristotelianSyllogism}.}
	\end{center}
\end{figure}

Set-Based interpretation, as we said, assumes that quantified statements and syllogisms are talking about relationships between sets. Both the knowledge representation of the propositions and the reasoning process can be represented through Venn Diagrams~\cite{Venn1880}. Let us consider again the example of Table~\ref{tab:AristotelianSyllogism}; Figure~\ref{fig:VennDiagramSyllogism} shows its representation using Venn Diagrams. Each one of the terms (\emph{Greeks}, \emph{human beings} and \emph{mortal beings}) is a set. Following the standard notation, the shadow areas denote the empty sets and the white ones those that have elements. As can be observed, the only white subset of \emph{Greeks} is also a subset of \emph{mortal beings}; i.e., ``all Greeks are mortal''.

In the literature, we can find several models for syllogistic reasoning that, assuming set-based interpretation and introducing new quantifiers, are compatible with the representation Venn Diagrams. Thus, Intermediate Syllogistics~\cite{Peterson2000,Thompson1986} introduces the quantifiers ``almost all'', ``few'', ``most'' and ``many'' (with the corresponding negations\footnote{The negation of ``almost all'' is ``few''.}; i.e, ``most\dots not'' and ``many\dots not'') as subdivisions into the space between the universal ``all'' and the existential ``some'' in the classical LSO (see Figure~\ref{fig:LSOClassical}). This allows the authors to generate a new LSO with six additional categorical statements and $81$ new inference patterns of syllogistic reasoning. Table~\ref{tab:IntermediateSyllogism} shows an example.

\begin{table}[h]%
	\centering
	\begin{tabular}{l}%
	  All students are tall\\
	  Most young people are students\\\hline
	  Most young people are tall
	\end{tabular}
	\caption{\label{tab:IntermediateSyllogism} Intermediate Syllogism.}
\end{table}

Generalized Intermediate Syllogistics~\cite{Murinova2012}, developed within the field of fuzzy logic, uses the so-called \emph{trichotomous evaluative linguistic} (TE\textsl{v}) expressions~\cite{Novak2008} to perform the subdivision between the universal and the existential quantifiers, but using fuzzy definitions instead of crisp ones, as in Intermediate Syllogistics. It is worth noting that all their syllogisms are proved as valid in this fuzzy version~\cite{Murinova2012}.\par

Other proposal is Exceptive Syllogistics~\cite{Murphree1991}. This reinterprets the quantifiers of the LSO in terms of exception ones, which have the form \emph{all but number}; e.g., ``all human beings are mortal'' is reinterpreted as ``all but $0$ human beings are mortal''. In general, ``all but $x$ S are P'', where $x$ denotes de size of the exception. This change generates a new perspective about the reasoning process, adopting an arithmetical character~\cite{Murphree1997}. Nevertheless, this entails to incorporate new presuppositions for dealing with the cardinality of $|S|$ and the corresponding subsets. For instance, we shall rewrite the argument of Table~\ref{tab:IntermediateSyllogism} in Table~\ref{tab:ExceptionSyllogism} using Exceptive Syllogistics approach. To obtain a result, we have to assume a cardinality for the set of students; e.g. $|students|=100$. As result we obtain ``all but $81$ young people are tall'', where ``all but $81$'' can be interpreted as ``most''.

\begin{table}[h]%
	\centering
	\begin{tabular}{l}%
	  All but $0$ students are tall\\
	  All but $19$ young people are students\\\hline
	  All but $100-19$ young people are tall
	\end{tabular}
	\caption{\label{tab:ExceptionSyllogism} Exceptive Syllogism.}
\end{table}

This way for modelling syllogisms demands more information or assumptions (as the size of $S$) than Intermediate Syllogistics, but it allows us to increase significantly the number of possible syllogisms to be addressed: the quantifier of the conclusion is directly calculated instead of determining the logical validity of each argument, as in the previous models.\par

Fuzzy Syllogistics~\cite{Zadeh1985} takes to its last consequences the arithmetical calculation of the conclusion of the syllogism and dismisses of the LSO. Thus, the intermediate regions between the universal and the existential quantifiers are denoted by fuzzy proportional numbers~\cite{Zadeh1983}. For instance, ``most students are tall'' means $\frac{|students \cap tall|}{|students|} \geq \alpha$, where $\alpha$ is any of the possible fuzzy numbers for interpreting ``most''. The reasoning process is based on the \emph{Quantifier Extension Principle} (QEP)~\cite{Zadeh1985} and is performed through fuzzy arithmetic. It is worth noting that some new inferences schemas are developed, as the example shown in Table~\ref{tab:FuzzySyllogism}, known as \emph{Intersection/product} syllogism. The quantifier of the conclusion is calculated using the fuzzy multiplication $\otimes$.

\begin{table}[h]%
	\centering
	\begin{tabular}{l}%
		Most students are young\\
		Most young students are single\\\hline
		$Most \otimes Most$ students are young and single
	\end{tabular}
	\caption{\label{tab:FuzzySyllogism} Fuzzy Syllogism.}
\end{table}

Interval Syllogistics~\cite{Dubois1988,Dubois1990,Dubois1993} follows the main ideas of Fuzzy Syllogistics, but avoiding the problems of fuzzy arithmetic that appear in reasoning~\cite{Liu1998}. In this case, each quantifier is associated to an interval or fuzzy set (defined through the Ker-Sup approach) and the reasoning process is not based on fuzzy arithmetic but on the maximization of intervals. For instance, ``most students are tall'' means $\frac{|students \cap tall|}{|students|}=[a,b]$ for intervals or $[a,c,d,b]$ for fuzzy sets, where the kernel is $Ker_{Q}=[c,d]$ and the support is $Sup_{Q}=[a,b]$.\par

It is worth noting that, although this framework is classified under the set-based interpretation, it is compatible with a probabilistic approach. In particular, with the so-called \emph{frequentism} interpretation of probability, advocated in 19th century by Jon Venn. He proposed that probabilities are identified with long-run frequencies of events~\cite{Korb2004} represented using Venn Diagrams. Thus, the concept of set is preserved as a collection of elements and quantifiers denote the frequencies associated to those set of elements; i.e., the main ideas of set-based interpretation. Like this, the reasoning process does not consist of a fuzzy arithmetic procedure but an optimization problem or, in probabilistic terms, restrictions propagation: the premises define the constraints to be satisfied and the conclusion the most favourable and most unfavourable proportions among the terms of the conclusion  according to premises. Table~\ref{tab:IntervalSyllogism} shows an example.

\begin{table}[h]%
\centering
	\begin{tabular}{l}%
		$[0.3,0.5]$ single people are young\\
		$[0.7,0.9]$ single people are students\\\hline
		$[0,0.5]$ single people are young and students
	\end{tabular}
	\caption{\label{tab:IntervalSyllogism} Interval Syllogism.}
\end{table}

\subsection{Some Open Questions}
\label{ssc:OpenQuestionsSetBased}
There are some questions inherited from classical syllogistics that must be addressed. Next, we shall analyse three: i) existential import, ii) singular statements and iii) non-proportional statements.

\subsubsection{Existential import}
\label{ssc:ExistentialImport}
A statement has existential content if it affirms that the terms of the proposition are not empty~\cite{Copi1994}. In ``some'' and ``some\dots not'' propositions, such as ``some animals are mammals'', this is clear, as their assertion necessarily entails the existence of elements in the involved sets. Nevertheless, in ``all'' and ``no'' statements this question is not so clear. For instance, let us consider the following example: ``all Martians are blond''. It means that every element of the set \emph{Martians} is also a member of the set \emph{blond people}. Today, there are not evidences of living beings in Mars, therefore, \emph{Martians} is a empty set. Since the empty set is subset of any set, this statement is true. Nonetheless, if we assume this interpretation, the subalternation relationship of the LSO is not satisfied, since ``all S are P'' entails ``some S are P'', which affirms that \emph{S} is not empty. This question is an old debate in philosophy and it comes from the Middle Ages. It is worth 
noting that existential import not only arises at universal statements but also at intermediate ones, such as the ones introduced by Intermediate Syllogistics (``almost all S are P'', ``most S are P'', ``many S are P'', ``few S are P'', etc.). We shall analyse some of them:
\begin{itemize}
 \item \textbf{almost all}: it seems clear that is closer to universal statements that existential ones, given the proper quantifier.
 \item \textbf{most}: we can find the following definition $most(S,P) \Leftrightarrow |S \cap P| > |S - P|$~\footnote{Typical definition for ``most'' interpreted  \emph{more than half of the} (on finite universes).} and, as we can see, it does not include existential import. Therefore, about this question, its behaviour is the same of universal statements.
 \item \textbf{many}: there is not an unique definition of this quantifier but it is defined analogously to ``most''.
 \item \textbf{few}: given that it is closer to ``some'', we can think that it can be seen as an addition to ``some'' preserving the existential import; however, its definition according to the Theory of Generalized Quantifiers~\cite{Barwise1981} does not include it.
\end{itemize}

In conclusion, we can see that the existential import problem takes more relevance it the expanded versions of the LSO as the new quantifiers are introduced as reductions of the universal statements.

A direct answer to this question is to assume that universal statements have existential import; i.e., the assertion of ``all Martians are blond'' entails that the set \emph{Martians} is not empty. However, this solution has several objections:
\begin{itemize}
 \item It is not compatible with the first order logic definition for universal statements; i.e, $\forall x (MARTIANx \rightarrow BLONDx)$.
 \item From the point of view of natural language, human beings do not assume existential import in any case and, therefore, this cannot directly incorporated to the meaning of the quantifiers~\cite{Kneale1968}.
\end{itemize}

\begin{figure}[h]
	\begin{center}
	\includegraphics[keepaspectratio=true,width=0.60\textwidth]{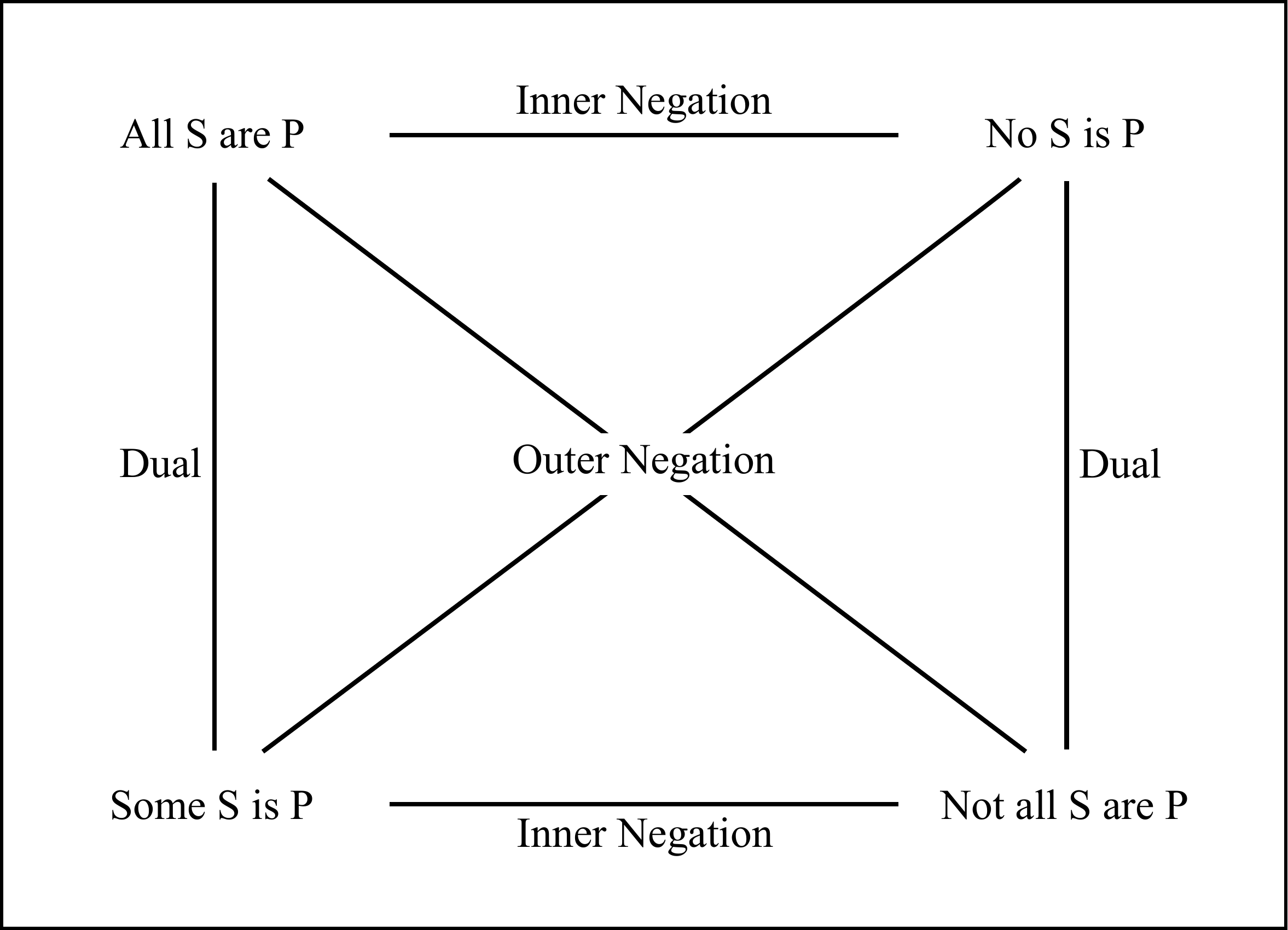}
	\caption{\label{fig:LSOModern} Modern version of the LSO without existential import.}
	\end{center}
\end{figure}

Other solution is to modify the LSO to obtain a new version without using subalternation property. This entails to use as core of the LSO the negation properties instead of the implication ones. Figure~\ref{fig:LSOModern}) shows the version of the LSO proposed by Peters and Westersth\r{a}lt~\cite{Peters2006}. As we can see, the subalternation property is not included in the schema and the negation relationships play the main role.\par

Notwithstanding, from the point of view of the reasoning process, to dispense with the existential import is not easy as many of the valid syllogisms use it. For instance, for those syllogistic models that entails arithmetic (such as Exceptive, Fuzzy or Interval Syllogistics), existential import allows us to avoid those undefinitions that can appear in the conclusion by $0$ at the denominator; those based on the LSO, there are several syllogistic models that cannot be proved; e.g., ``all DT are MT, all NT are DT; therefore, some NT are MT''.\par 

Thus, we propose to introduce the existential import as an explicit assumption when it is necessary. Then, any syllogism must include as additional explicit presupposition a statement that says that all the subject-terms of the premises are non-empty sets; for instance, ``no consulting detective is Spanish; there is at least one consulting detective''. For everyday reasoning, this is not so strange as human beings perform a lot of additional assumptions or presuppositions. Existential import is only one more that should be explicitly uttered to avoid misunderstanding, as it is not always assumed by anybody.

\subsubsection{Singular Statements}
\label{ssc:SingularStatements}
A syllogism, in its standard definition as we said, only deals with categorical statements with the form \emph{Q S are P}. However, there are other type of statements, such as ``Socrates is a human being'', where the subject-term is an individual instead of a set or a category. These are the so-called \emph{singular statements}~\cite{Kneale1968}. They are very common in natural language and in everyday reasoning. For instance, let us consider the syllogism of Table~\ref{tab:SingularSyllogism}, where the second premise is a singular statement. If we apply the standard definition of syllogism, it is not an example one; however, anybody accept its validity and it also involves quantified statements (the first premise).

\begin{table}[h]%
	\centering
	\begin{tabular}{l}%
		All human beings are mortal\\
		Socrates is a human being\\\hline
		Socrates is mortal
	\end{tabular}
	\caption{\label{tab:SingularSyllogism} Syllogism involving a singular statement.}
\end{table}

This question can be easily addressed using the concept of \emph{singleton}; i.e., a set with exactly one element. Thus, any singular statement denotes a single element in the subject-term; i.e., ``Socrates'' denotes the set $\{Socrates\}$, whose unique element is the historical personage \emph{Socrates}. This leads us to consider the proper name \emph{Socrates} as a \emph{defined description}~\cite{Russell1905}. Therefore, any singular statement that appears in a syllogism involving singular statements can be transformed into a categorical one transforming the defined description into a quantified formula with ``all''; i.e., ``Socrates $\equiv$ all Socrates''. 

\subsubsection{Non-Proportional Statements}
\label{sssc:NonProportionalSet}
As we show, the Aristotelian and most of the current approaches to syllogistic reasoning only considers, in general, proportional quantifiers, such as ``all'', ``few'', ``most'', ``many'', etc. However, we manage in natural language many more (for instance, ``all but three'', ``double'', ``half'', ``twenty-five'', etc.) that are not usually considered in syllogistic models.\par

The typical fuzzy classification of quantifiers~\cite{Zadeh1984} only considers two types of binary quantifiers: absolute ones (such as ``four'', ``around twenty-five'', etc.) and proportional ones (i.e., ``few'', ``most'', ``many'', etc.). Nevertheless, the current linguistic theory of quantification, the Theory of Generalized Quantifiers (TGQ)~\cite{Barwise1981} distinguishes many more types of quantifiers, as we pointed out. In addition of the absolute and proportional types, there are others, such as the pointed out exception ones (i.e., ``all but three'', ``all but around five'', etc.) or the comparative ones (i.e., ``double'', ``half'', etc.). Thus, from our point of view, any further improvement in the modelling of syllogistic reasoning should include these types of quantifiers.

TGQ defines the quantification phenomenon in terms of relationships between sets. This means that TGQ and set-based interpretation are fully compatible, which provide us a framework for dealing with non-proportional quantifiers. The standard notation for Venn Diarams is not enough for dealing with this type of statements, but using set-theory notation, it is not difficult. For instance, let us consider an universe $E$ and the exception statement ``All but three $Y_{1}$ are $Y_{2}$''; for $Y_{1}, Y_{2} \in \mathcal{P}(E)$:

\small
\begin{eqnarray*}
\text{\textit{All but three}}: \mathcal{P}(E) \times \mathcal{P}(E) &\rightarrow& \left\{0,1\right\}\\
																	(Y_{1},Y_{2}) &\rightarrow& \text{\textit{All but three}}(Y_{1},Y_{2})= \left\{ 
																															  \begin{aligned}
																												          & 0: if &|Y_{1} \cap \overline{Y_{2}}| \neq 3\\
																															   	& 1: if &|Y_{1} \cap \overline{Y_{2}}| = 3
																												        \end{aligned} \right.
\end{eqnarray*}

\normalsize
In~\cite{Oaksford2007}, this approach is criticised as it entails to apply high order logic. As it is known, high order logics are not completely decidable and, then, there are propositions that cannot be proved or rejected. However, there are many statements that are decidable and, very likely, the formalization of most of quantified statements that we use in our daily life belong to this class. Thus, we consider that the compatibility with TGQ is meaningfully step in the improvement of syllogistic reasoning.

\section{Conditional Interpretation: Models and Questions}
\label{sc:ConditionalInterpretation}
Conditional interpretation, as we said, assumes that syllogistic statements have a conditional deep form with the following equivalence, ``if $x$ \emph{is S}, then $x$ \emph{is P} with link \emph{Q} between them'', where \emph{S} is the subject-term, \emph{P} the predicate-term and \emph{Q} the indicates how strong is the link between \emph{S} and \emph{P}. In this case, it is relevant to note that propositions do not involve sets or relationships between sets but individuals and properties in an unconditional universe. For instance, the statement ``all human beings are mortal'' means that any element $x$ in the universe $U$, if $x$ has the property of being \emph{a human being}, then $x$ has the property of being \emph{a mortal being} with a very strong degree of connection denoted by ``all''. If the quantifier is other (non-universal), such as ``almost all'', ``few'', etc. this degree is going weaker.\par

In probability terms, this interpretation is closer to think probability as reporting our subjective degrees of belief about a fact or an event. This is the view expressed by Thomas Bayes and P. S. Laplace~\cite{Korb2004}. Thus, under this perspective, reasoning process can be viewed as an heuristic process where a subject calculate the probability of the conclusion according to the probabilities expressed in the premises.\par

There are different proposals for modelling the strength of the conditional (~\cite{Oaksford2007,Spies1989,Schwartz1997}), but in this paper we only focus on the so-called Probabilistic Syllogistics~\cite{Oaksford2007}, as we think that is is the most relevant one.\par

Probabilistic Syllogistics is developed into the field of psychology of reasoning and it focus on analysing how human beings reason rather than which are the rules of the right reasoning~\cite{Oaksford2007}, which reinforces the subjective interpretation of probability. The main thesis of the authors is that human mind is probabilistic; therefore, quantifiers are a linguistic way for expressing probabilities. For instance, ``most S are P'' means $1 - \epsilon \leq P(P|S) < 1, \epsilon > 0$, where $\epsilon$ is small as it represents the meaning of ``few'', the antonym of ``most''. Table~\ref{tab:ProbabilisticReadingQuantifiers} shows the probabilistic notation of the four Aristotelian quantifiers, ``most'' and ``few''.

\begin{table}[h]
	\centering
		\begin{tabular}{ll}
		\multicolumn{1}{c}{\textbf{Linguistic Expression}}	&	\multicolumn{1}{c}{\textbf{Probabilistic Expression}}\\
		\hline
		All S are P						& $P(P|S)=1$\\
		No S are P						& $P(P|S)=0$\\
		Some S are P						& $P(P|S)>0$ and $S$ are not empty\\
		Some S are not P					& $P(P|S)<1$ and $S$ are not empty\\
		Most S are P						& $1 - \epsilon \leq P(P|S) < 1, \epsilon > 0$\\
		Few S are P						& $0 < P(P|S) \leq \epsilon, \epsilon > 0$\\
		\hline
		\end{tabular}
	\caption{\label{tab:ProbabilisticReadingQuantifiers} Probabilistic expression of ``all'', ``none'', ``some'', ``some\dots not'', ``few'' and ``most''.}

\end{table}

Regarding inference, the rules that work in this model are heuristic rather than deductive, as they are focused on searching the most plausible solution, putting the logical validity of the argument in a side place. Let us consider the example shown in Table~\ref{tab:ProbabilisticInferenceFigureI}. The premises are organized according to their quantifiers; i.e., from the universal one (\emph{all}-premises) in the first premise to existential one in the second one (\emph{some}-premises). With respect to the conclusion, two of the heuristic rules appears~\cite[p. 217-219]{Oaksford2007}: 
\begin{itemize}
 \item \emph{min}-heuristic rule: it indicates that there is a tendency towards selecting the quantifier of the least informative premise (\emph{some} and \emph{some\dots not}-premises.).
 \item \emph{attachment}-heuristic rule: it indicates that there is a tendency towards selecting the subject of one of the premises as the subject of the conclusion.
\end{itemize}

\begin{table}[h]
	\centering
	\begin{tabular}{ll}
		\multicolumn{1}{c}{\textbf{Syllogism}}					&\multicolumn{1}{c}{\textbf{Heuristic}}\\
		\hline
		All students are tall 						&\emph{max}-premise\\
		Some young people are students					&\emph{min}-premise\\\cline{1-1}
		\emph{Some}-type conclusion	& by \emph{min}-heuristic\\
		Some young people are students					& by \emph{attachment}-heuristic\\
	\end{tabular}
\caption{Inference schema of syllogism AII (Figure I) under a probabilistic approach.}
\label{tab:ProbabilisticInferenceFigureI}
\end{table}

It is clear that this way for interpreting syllogisms is also deductive, but it does no have the probative character of the set-based models as this focuses on finding a quick (although seemingly right) answer.

\subsection{Some Open Questions}
\label{ssc:OpenQuestionsConditional}
Next, we shall consider the three same questions analysed in section~\ref{ssc:OpenQuestionsSetBased}: i) the existential import, ii) singular statements and iii) non-proportional quantifiers.\par

\subsubsection{Existential Import}
As can be observed at Table~\ref{tab:ProbabilisticReadingQuantifiers}, existential import only appears in ``some'' and ``some\dots not'' statements; the remaining ones, universal, ``most'' and ``few'' do not include it. In this case, given there is not LSO there is not any problem with subalternation problem. On the other hand, this approach follows the same idea of the modern LSO described in Figure~\ref{fig:LSOModern}.\par

As we explained in section~\ref{sc:ConditionalInterpretation}, it is assumed that these statements deal with individuals and properties. Therefore, the concept of empty set has no place here; a concrete individual can have a property or not and this determine the truth value of the corresponding part of the conditional. For instance, if we use the material conditional, ``if $x$ is a Martian, then $x$ is blond with maximum strength in the connection'' is vacuously true because no $x$ has the the property of being a \emph{Martian}. From the point of view of the heuristic rules defined by Probabilistic Syllogistics, the existential import does not have any impact as there are neither rules about subalternation not moods analogous to the Aristotelian ones.

\subsubsection{Singular Statements}
In the conditional interpretation, it is not a clear definition for singular statements in terms of probability. For instance, let us consider the proposition ``Socrates is mortal''. We can consider its conditional definition in terms of ``if $x$ is Socrates, then $x$ is mortal''. However, which is the quantifier that defines the strength of the link between the antecedent and the consequent? If \emph{Socrates} is a property, it cannot be a quantifier.\par

A possible solution is to assume that the predication of the property \emph{Socrates} entails the existential quantifier; i.e., ``Socrates is mortal'' is equivalent to ``if $x$ is Socrates, then $x$ is mortal and there is an $x$ that is Socrates''. However, as it was explained in Table~\ref{tab:ProbabilisticReadingQuantifiers}, ``some'' quantifier does not denote a strong quantifier between antecedent and consequent which seems counterintuitive with respect to this case, where only a single element has the property of being Socrates. On the other hand, this lead us to the problem of the existential import. For instance, let us consider the following statement ``the present king of France is bold''. We know that France is a republic and it dos not have king, then, the predicate term of this statement has no reference. Thus, is this proposition true or false? This is a debatable question in philosophy of language.

An alternative possibility is explaining singular statements using probabilistic terms. These can be interpreted as evidences~\cite{Korb2004}, new pieces of information that allows us to update our beliefs and, therefore, accepted as true in the model. This updating process is can bee seen as the equivalent to reasoning, and the usual form for expressing it  is trough \emph{Modus Pones}; i.e., ``if A is B, $x$ is A then $x$ is B''. However, in this type of arguments, the role of the quantifier, as a linguistic particle for denoting quantity, is hidden.

\subsubsection{Non-proportional quantifiers}
\label{sssc:NonProportionalConditional}
This question was explicitly addressed in~\cite{Oaksford2007}. The authors defend that this question is not relevant enough. The use of precise absolute quantifiers is very reduced in our daily life as they only appear in very specific contexts. In these cases, precise assertions can be interpreted in terms of proportional ones if the size of the cardinality of the subject-term is known or directly estimated.\par

On the other hand, the role of the TGQ for dealing with a reasoning model also was questioned. It is based on a second order logic and this type of logics, in general, is not completely decidable and, therefore, there are propositions that cannot be proved or rejected. There is an exception with monadic second-order logic, which is decidable but it can only express propositions composed by a single argument such as ``Socrates is mortal''. However, most of the quantified statements managed by human being are binary (i.e., ``few students are tall'') and some of them cannot be decidable. Lastly, conditional interpretation is better supported by psychological experiments than set-based interpretation. Therefore, other quantifiers that appear in natural language such as the absolute or exception ones are not relevant for this model of syllogism.\par

Notwithstanding, this is not a minor question from our point of view whether a computational perspective is adopted. In~\cite{Westersthal1989}, the compatibility between set-based interpretation and TGQ was proved. TGQ is not only a logical theory but a linguistic one; currently, it is the prevailing theory of quantification in natural language.\par

On the other hand, the use of non-proportional quantifiers can be relevant in many contexts where the everyday reasoning is not applied but natural language is used, such as diagnostic system. Therefore, for a further improvement of syllogistic reasoning, this type of quantifiers must not be dismissed.

\section{Conclusions}
\label{sc:Conclusions}
We have presented two alternative interpretations for quantified statements in syllogisms: one based on the management of sets and the other on individuals and properties involved in conditional statements. This entails not only a change in the knowledge representation (from sets of elements to assigning properties to individuals) but also in the reasoning process (from one that tries to prove without doubt to other that tries to obtain quick answers).\par

Three relevant questions of syllogistic reasoning have been analysed (existential import, singular statements and non-proportional quantifiers) and how they can be addressed from both models. We conclude that set-based interpretation shows, in general, a better behaviour for a further development of syllogistic reasoning. In the vein of improving the expressiveness capability of syllogistic models, the analysis of non-proportional quantifiers is more relevant than the other two questions, as it allows us to deal with many of the quantifiers defined by TGQ. In this way, many other arguments of everyday reasoning can be addressed using a logical model.

\section*{Acknowledgments}
This work was supported in part by Spanish Ministry for Economy and Innovation and the European Regional Development Fund (ERDF/ FEDER) (grant TIN2011-29827-C02-02) and by the European Regional Development Fund \newline (ERDF/ FEDER) under the projects CN2012/151 and CN2011/058 of the Galician Ministry of Education. I also wish to thank Alejandro Sobrino and anonymous referees and the editors for their useful and constructive comments and insightful suggestions on the previous versions of this paper that allowed us to highly improve its contents.

\bigskip

\bibliographystyle{aphsc_article}

\footnotesize

\bibliography{mybib}


\begin{thebibliography}{10}
\providecommand{\url}[1]{[#1]}
\providecommand{\urlprefix}{}

\bibitem{Aristotle1949}
Aristotle: \emph{Prior and posterior analytics}. Oxford: Clarendom Press 1949.

\bibitem{Kneale1968}
Kneale W, Kneale M: \emph{The Development of Logic}. Oxford: Clarendon Press
  1968.

\bibitem{Lukasiewicz1957}
Lukasiewicz J: \emph{Aristotle's Syllogistic}. Oxford: The Clarendon Press
  1957.

\bibitem{Peterson2000}
Peterson PL: \emph{Intermediate {Q}uantities. {L}ogic, linguistics, and
  {A}ristotelian semantics}. Alsershot. England: Ashgate 2000.

\bibitem{Thompson1982}
Thompson B: \textbf{Syllogisms Using ``Few'', ``Many'', and ``Most''}.
  \emph{Notre Dame Journal of Formal Logic} 1982, \textbf{23}:75--84.

\bibitem{Murphree1997}
Murphree WA: \textbf{The Numerical Syllogism and Existential Presupposition}.
  \emph{Notre Dame Journal of Formal Logic} 1997, \textbf{38}:49 -- 64.

\bibitem{Murinova2012}
Murinov\'a P, Nov\'ak V: \textbf{A formal theory of generalized intermediate
  syllogisms}. \emph{Fuzzy Sets and Systems} 2012, \textbf{186}:47--80.

\bibitem{Zadeh1985}
Zadeh LA: \textbf{Syllogistic reasoning in fuzzy logic and its applications to
  usuality and reasoning with dispositions}. \emph{IEEE Transactions On
  Systems, Man and Cybernetics} 1985, \textbf{15}(6):754--765.

\bibitem{Dubois1990}
Dubois D, Prade H, Toucas JM: \emph{Intelligent Systems. State of the Art and
  Future Directions}, Great Britain: Ellis Horwood 1990 chap. Inference with
  imprecise numerical quantifiers, :52--72.

\bibitem{Oaksford2007}
Oaksford M, Chater N: \emph{Bayesian Rationality. The probabilistic approach to
  human reasoning}. Oxford: Oxford University Press 2007.

\bibitem{Spies1989}
Spies M: \emph{Syllogistic Inference under Uncertainty}. Psychologie Verlags
  Union 1989.

\bibitem{Pereira2013}
Pereira-Fari{\~n}a M, D\'iaz-Hermida F, Bugar\'in A: \textbf{On the analysis of
  set-based fuzzy quantified reasoning using classical syllogistics}.
  \emph{Fuzzy Sets and Systems} 2013, \textbf{214}:83 -- 94.

\bibitem{Venn1880}
Venn J: \textbf{On the Diagrammatic and Mechanical Representation of
  Propositions and Reasonings}. \emph{Philosophical Magazine and Journal of
  Science} 1880, \textbf{9}(59):1--18.

\bibitem{Thompson1986}
Thompson B: \textbf{Syllogisms with Statistical Quantifiers}. \emph{Notre Dame
  Journal of Formal Logic} 1986, \textbf{27}:93--103.

\bibitem{Novak2008}
Nov\'ak V: \textbf{A formal theory of intermediate quantifiers}. \emph{Fuzzy
  Sets and Systems} 2008, \textbf{159}:1229--1246.

\bibitem{Murphree1991}
Murphree WA: \emph{Numerically Exceptive Logic: A Reduction of the Classical
  Syllogism}. New York: Peter Lang 1991.

\bibitem{Zadeh1983}
Zadeh LA: \textbf{A {C}omputational {A}pproach to {F}uzzy {Q}uantifiers in
  {N}atural {L}anguage}. \emph{Computer and Mathematics with Applications}
  1983, \textbf{8}:149--184.

\bibitem{Dubois1988}
Dubois D, Prade H: \textbf{On fuzzy syllogisms}. \emph{Computational
  Intelligence} 1988, \textbf{4}(2):171--179.

\bibitem{Dubois1993}
Dubois D, Godo L, L\'opez~de M\'antaras R, Prade H: \textbf{Qualitative
  {R}easoning with {I}mprecise {P}robabilities}. \emph{Journal of Intelligent
  Information Systems} 1993, \textbf{2}:319--363.

\bibitem{Liu1998}
Liu Y, Kerre EE: \textbf{An overview of fuzzy quantifiers (II). Reasoning and
  applications}. \emph{Fuzzy Sets and Systems} 1998, \textbf{95}:135--146.

\bibitem{Korb2004}
Korb K, Nicholson A: \emph{Bayesian Artificial Intelligence}. Chapman \&
  Hall/CRC 2004.

\bibitem{Copi1994}
Copi IM: \emph{Introduction to logic}. New York: MacMillan 1994.

\bibitem{Barwise1981}
Barwise J, Cooper R: \textbf{Generalized quantifiers and natural language}.
  \emph{Linguistics and Philosophy} 1981, \textbf{4}:159--219.

\bibitem{Peters2006}
Peters S, Westersth\aa{}l D: \emph{Quantifiers in language and logic}. Oxford:
  Clarendom Press 2006.

\bibitem{Russell1905}
Russell B: \textbf{On Denoting}. \emph{Mind} 1905, \textbf{14}(56):479--493,
  \urlprefix\url{[http://www.jstor.org/stable/2248381]}.

\bibitem{Zadeh1984}
Zadeh LA: \emph{Aspects of {V}agueness}, Dodrecht: D. Reidel 1984 chap. A
  Theory of Commonsense Knowledge, :257--296.

\bibitem{Schwartz1997}
Schwartz DG: \textbf{Dynamic reasoning with qualified syllogisms}.
  \emph{Artifical Intelligence} 1997, \textbf{93}:103--167.

\bibitem{Westersthal1989}
Westersth\aa{}l D: \textbf{Aristotelian Syllogisms and Generalized
  Quantifiers}. \emph{Studia Logica} 1989, \textbf{48}(4):577--585.

\end{thebibliography}

\newcommand{\BMCxmlcomment}[1]{}

\BMCxmlcomment{

<refgrp>

<bibl id="B1">
  <title><p>Prior and posterior analytics</p></title>
  <aug>
    <au><cnm>Aristotle</cnm></au>
  </aug>
  <publisher>Oxford: Clarendom Press</publisher>
  <pubdate>1949</pubdate>
</bibl>

<bibl id="B2">
  <title><p>The Development of Logic</p></title>
  <aug>
    <au><snm>Kneale</snm><fnm>W.</fnm></au>
    <au><snm>Kneale</snm><fnm>M.</fnm></au>
  </aug>
  <publisher>Oxford: Clarendon Press</publisher>
  <pubdate>1968</pubdate>
</bibl>

<bibl id="B3">
  <title><p>Aristotle's Syllogistic</p></title>
  <aug>
    <au><snm>Lukasiewicz</snm><fnm>J.</fnm></au>
  </aug>
  <publisher>Oxford: The Clarendon Press</publisher>
  <pubdate>1957</pubdate>
</bibl>

<bibl id="B4">
  <title><p>Intermediate {Q}uantities. {L}ogic, linguistics, and {A}ristotelian
  semantics</p></title>
  <aug>
    <au><snm>Peterson</snm><fnm>P. L.</fnm></au>
  </aug>
  <publisher>Alsershot. England: Ashgate</publisher>
  <pubdate>2000</pubdate>
</bibl>

<bibl id="B5">
  <title><p>Syllogisms Using ``Few'', ``Many'', and ``Most''</p></title>
  <aug>
    <au><snm>Thompson</snm><fnm>B.</fnm></au>
  </aug>
  <source>Notre Dame Journal of Formal Logic</source>
  <pubdate>1982</pubdate>
  <volume>23</volume>
  <issue>1</issue>
  <fpage>75</fpage>
  <lpage>-84</lpage>
</bibl>

<bibl id="B6">
  <title><p>The Numerical Syllogism and Existential Presupposition</p></title>
  <aug>
    <au><snm>Murphree</snm><fnm>WA</fnm></au>
  </aug>
  <source>Notre Dame Journal of Formal Logic</source>
  <pubdate>1997</pubdate>
  <volume>38</volume>
  <issue>1</issue>
  <fpage>49</fpage>
  <lpage>64</lpage>
</bibl>

<bibl id="B7">
  <title><p>A formal theory of generalized intermediate syllogisms</p></title>
  <aug>
    <au><snm>Murinov\'a</snm><fnm>P.</fnm></au>
    <au><snm>Nov\'ak</snm><fnm>V.</fnm></au>
  </aug>
  <source>Fuzzy Sets and Systems</source>
  <pubdate>2012</pubdate>
  <volume>186</volume>
  <fpage>47</fpage>
  <lpage>80</lpage>
</bibl>

<bibl id="B8">
  <title><p>Syllogistic reasoning in fuzzy logic and its applications to
  usuality and reasoning with dispositions</p></title>
  <aug>
    <au><snm>Zadeh</snm><fnm>L. A.</fnm></au>
  </aug>
  <source>IEEE Transactions On Systems, Man and Cybernetics</source>
  <pubdate>1985</pubdate>
  <volume>15</volume>
  <issue>6</issue>
  <fpage>754</fpage>
  <lpage>-765</lpage>
</bibl>

<bibl id="B9">
  <title><p>Intelligent Systems. State of the Art and Future
  Directions</p></title>
  <aug>
    <au><snm>Dubois</snm><fnm>D.</fnm></au>
    <au><snm>Prade</snm><fnm>H.</fnm></au>
    <au><snm>Toucas</snm><fnm>J M.</fnm></au>
  </aug>
  <publisher>Great Britain: Ellis Horwood</publisher>
  <editor>Z. W. Ras M. Remankova</editor>
  <section><title><p>Inference with imprecise numerical
  quantifiers</p></title></section>
  <pubdate>1990</pubdate>
  <fpage>52</fpage>
  <lpage>72</lpage>
</bibl>

<bibl id="B10">
  <title><p>Bayesian Rationality. The probabilistic approach to human
  reasoning</p></title>
  <aug>
    <au><snm>Oaksford</snm><fnm>M.</fnm></au>
    <au><snm>Chater</snm><fnm>N.</fnm></au>
  </aug>
  <publisher>Oxford: Oxford University Press</publisher>
  <pubdate>2007</pubdate>
</bibl>

<bibl id="B11">
  <title><p>Syllogistic Inference under Uncertainty</p></title>
  <aug>
    <au><snm>Spies</snm><fnm>M.</fnm></au>
  </aug>
  <publisher>Psychologie Verlags Union</publisher>
  <pubdate>1989</pubdate>
</bibl>

<bibl id="B12">
  <title><p>On the analysis of set-based fuzzy quantified reasoning using
  classical syllogistics</p></title>
  <aug>
    <au><snm>Pereira Fari{\~n}a</snm><fnm>M.</fnm></au>
    <au><snm>D\'iaz Hermida</snm><fnm>F.</fnm></au>
    <au><snm>Bugar\'in</snm><fnm>A.</fnm></au>
  </aug>
  <source>Fuzzy Sets and Systems</source>
  <pubdate>2013</pubdate>
  <volume>214</volume>
  <fpage>83</fpage>
  <lpage>94</lpage>
</bibl>

<bibl id="B13">
  <title><p>On the Diagrammatic and Mechanical Representation of Propositions
  and Reasonings</p></title>
  <aug>
    <au><snm>Venn</snm><fnm>J.</fnm></au>
  </aug>
  <source>Philosophical Magazine and Journal of Science</source>
  <pubdate>1880</pubdate>
  <volume>9</volume>
  <issue>59</issue>
  <fpage>1</fpage>
  <lpage>18</lpage>
</bibl>

<bibl id="B14">
  <title><p>Syllogisms with Statistical Quantifiers</p></title>
  <aug>
    <au><snm>Thompson</snm><fnm>B.</fnm></au>
  </aug>
  <source>Notre Dame Journal of Formal Logic</source>
  <pubdate>1986</pubdate>
  <volume>27</volume>
  <issue>1</issue>
  <fpage>93</fpage>
  <lpage>-103</lpage>
</bibl>

<bibl id="B15">
  <title><p>A formal theory of intermediate quantifiers</p></title>
  <aug>
    <au><snm>Nov\'ak</snm><fnm>V.</fnm></au>
  </aug>
  <source>Fuzzy Sets and Systems</source>
  <pubdate>2008</pubdate>
  <volume>159</volume>
  <fpage>1229</fpage>
  <lpage>1246</lpage>
</bibl>

<bibl id="B16">
  <title><p>Numerically Exceptive Logic: A Reduction of the Classical
  Syllogism</p></title>
  <aug>
    <au><snm>Murphree</snm><fnm>WA</fnm></au>
  </aug>
  <publisher>New York: Peter Lang</publisher>
  <pubdate>1991</pubdate>
</bibl>

<bibl id="B17">
  <title><p>A {C}omputational {A}pproach to {F}uzzy {Q}uantifiers in {N}atural
  {L}anguage</p></title>
  <aug>
    <au><snm>Zadeh</snm><fnm>L. A.</fnm></au>
  </aug>
  <source>Computer and Mathematics with Applications</source>
  <pubdate>1983</pubdate>
  <volume>8</volume>
  <fpage>149</fpage>
  <lpage>-184</lpage>
</bibl>

<bibl id="B18">
  <title><p>On fuzzy syllogisms</p></title>
  <aug>
    <au><snm>Dubois</snm><fnm>D.</fnm></au>
    <au><snm>Prade</snm><fnm>H.</fnm></au>
  </aug>
  <source>Computational Intelligence</source>
  <pubdate>1988</pubdate>
  <volume>4</volume>
  <issue>2</issue>
  <fpage>171</fpage>
  <lpage>-179</lpage>
</bibl>

<bibl id="B19">
  <title><p>Qualitative {R}easoning with {I}mprecise
  {P}robabilities</p></title>
  <aug>
    <au><snm>Dubois</snm><fnm>D.</fnm></au>
    <au><snm>Godo</snm><fnm>L.</fnm></au>
    <au><snm>M\'antaras</snm><fnm>R.</fnm></au>
    <au><snm>Prade</snm><fnm>H.</fnm></au>
  </aug>
  <source>Journal of Intelligent Information Systems</source>
  <pubdate>1993</pubdate>
  <volume>2</volume>
  <fpage>319</fpage>
  <lpage>-363</lpage>
</bibl>

<bibl id="B20">
  <title><p>An overview of fuzzy quantifiers (II). Reasoning and
  applications</p></title>
  <aug>
    <au><snm>Liu</snm><fnm>Y.</fnm></au>
    <au><snm>Kerre</snm><fnm>E. E.</fnm></au>
  </aug>
  <source>Fuzzy Sets and Systems</source>
  <pubdate>1998</pubdate>
  <volume>95</volume>
  <fpage>135</fpage>
  <lpage>-146</lpage>
</bibl>

<bibl id="B21">
  <title><p>Bayesian Artificial Intelligence</p></title>
  <aug>
    <au><snm>Korb</snm><fnm>K.B.</fnm></au>
    <au><snm>Nicholson</snm><fnm>A.E.</fnm></au>
  </aug>
  <publisher>Chapman \& Hall/CRC</publisher>
  <pubdate>2004</pubdate>
</bibl>

<bibl id="B22">
  <title><p>Introduction to logic</p></title>
  <aug>
    <au><snm>Copi</snm><fnm>IM</fnm></au>
  </aug>
  <publisher>New York: MacMillan</publisher>
  <pubdate>1994</pubdate>
</bibl>

<bibl id="B23">
  <title><p>Generalized quantifiers and natural language</p></title>
  <aug>
    <au><snm>Barwise</snm><fnm>J.</fnm></au>
    <au><snm>Cooper</snm><fnm>R.</fnm></au>
  </aug>
  <source>Linguistics and Philosophy</source>
  <pubdate>1981</pubdate>
  <volume>4</volume>
  <fpage>159</fpage>
  <lpage>-219</lpage>
</bibl>

<bibl id="B24">
  <title><p>Quantifiers in language and logic</p></title>
  <aug>
    <au><snm>Peters</snm><fnm>S.</fnm></au>
    <au><snm>Westersth\aa{}l</snm><fnm>D.</fnm></au>
  </aug>
  <publisher>Oxford: Clarendom Press</publisher>
  <pubdate>2006</pubdate>
</bibl>

<bibl id="B25">
  <title><p>On Denoting</p></title>
  <aug>
    <au><snm>Russell</snm><fnm>B</fnm></au>
  </aug>
  <source>Mind</source>
  <publisher>Oxford University Press on behalf of the Mind
  Association</publisher>
  <series><title><p>New Series</p></title></series>
  <pubdate>1905</pubdate>
  <volume>14</volume>
  <issue>56</issue>
  <fpage>479</fpage>
  <lpage>493</lpage>
  <url>http://www.jstor.org/stable/2248381</url>
</bibl>

<bibl id="B26">
  <title><p>Aspects of {V}agueness</p></title>
  <aug>
    <au><snm>Zadeh</snm><fnm>L. A.</fnm></au>
  </aug>
  <source>Aspects of Vagueness</source>
  <publisher>Dodrecht: D. Reidel</publisher>
  <editor>Skala, H. J. and Termini, S. and Trillas, E.</editor>
  <section><title><p>A Theory of Commonsense Knowledge</p></title></section>
  <pubdate>1984</pubdate>
  <fpage>257</fpage>
  <lpage>296</lpage>
</bibl>

<bibl id="B27">
  <title><p>Dynamic reasoning with qualified syllogisms</p></title>
  <aug>
    <au><snm>Schwartz</snm><fnm>DG</fnm></au>
  </aug>
  <source>Artifical Intelligence</source>
  <pubdate>1997</pubdate>
  <volume>93</volume>
  <fpage>103</fpage>
  <lpage>167</lpage>
</bibl>

<bibl id="B28">
  <title><p>Aristotelian Syllogisms and Generalized Quantifiers</p></title>
  <aug>
    <au><snm>Westersth\aa{}l</snm><fnm>D.</fnm></au>
  </aug>
  <source>Studia Logica</source>
  <pubdate>1989</pubdate>
  <volume>48</volume>
  <issue>4</issue>
  <fpage>577</fpage>
  <lpage>-585</lpage>
</bibl>

</refgrp>
} 

\end{document}